\documentclass{article}

\usepackage[final]{corl_2023} 
\usepackage{float}
\usepackage{subcaption}
\usepackage{graphicx}
\usepackage{amsmath}

\title{Curiosity \& Entropy Driven Unsupervised RL in Multiple Environments}

\author{
  Shaurya Dewan\\
  Robotics Institute \\ 
  Carnegie Mellon University \\
  \texttt{srdewan@andrew.cmu.edu} \\
  \And
   Anisha Jain \\
   Robotics Institute \\ 
  Carnegie Mellon University \\
  \texttt{anishaja@andrew.cmu.edu} \\
  \AND
   Zo\"e LaLena \\
   Robotics Institute \\ 
  Carnegie Mellon University \\
  \texttt{zlalena@andrew.cmu.edu} \\
  \And
  Lifan Yu \\
  Robotics Institute \\ 
  Carnegie Mellon University \\
  \texttt{lifany@andrew.cmu.edu} \\
}

\begin{document}
\maketitle


\begin{abstract}
    The authors of \cite{unsup} propose a method, $\alpha$MEPOL, to tackle unsupervised RL across multiple environments. They pre-train a task-agnostic exploration policy using interactions from an entire environment class and then fine-tune this policy for various tasks using supervision. We expanded upon this work, with the goal of improving performance. We primarily propose and experiment with five new modifications to the original work: sampling trajectories using an entropy-based probability distribution, dynamic alpha, higher KL Divergence threshold, curiosity-driven exploration, and $\alpha$-percentile sampling on curiosity. Dynamic alpha and higher KL-Divergence threshold both provided a significant improvement over the baseline from [1]. PDF-sampling failed to provide any improvement due to it being approximately equivalent to the baseline method when the sample space is small.  In high-dimensional environments, the addition of curiosity-driven exploration enhances learning by encouraging the agent to seek diverse experiences and explore the unknown more. However, its benefits are limited in low-dimensional and simpler environments where exploration possibilities are constrained and there is little that is truly unknown to the agent. Overall, some of our experiments did boost performance over the baseline and there are a few directions that seem promising for further research.
\end{abstract}



\section{Introduction}
Unsupervised Reinforcement Learning (RL) is vital for advancing autonomous systems, enabling agents to learn from unannotated data in real-world scenarios where labeled datasets are scarce. This approach facilitates autonomous exploration and adaptation to environments without explicit guidance, fostering adaptability and scalability.

In diverse, non-overlapping environments, unsupervised RL becomes essential, allowing agents to acquire adaptable policies for seamless transitions between tasks. Unlike traditional approaches relying on fine-tuning with pre-trained policies, there's a lack of consensus on the optimal pre-training objective.

The Maximum State Visitation Entropy (MSVE) objective offers an effective alternative, encouraging agents to learn policies maximizing state visitation entropy. Despite the success of unsupervised pre-training in reducing reliance on reward functions, most solutions assume a single-environment scenario, limiting their applicability to diverse and dynamic settings.

In this work, we aim to build upon the work of \cite{unsup}, to devise a novel unsupervised RL framework that seamlessly integrates curiosity-driven exploration and state visitation entropy maximization across a domain of environments. The purpose of the proposed approach is to develop a more robust and efficient RL agent capable of exploring and learning in complex and diverse environments.

\section {Previous Work}
In this section, we review previous works that try to tackle the problem of unsupervised RL in multiple environments. 

\cite{AgentPractice} previously explored a learning process involving agnostic pre-training (referred to as "practice") and supervised fine-tuning ("match") in a specific class of environments. However, they alternated these phases, using match supervision to learn the reward for practice through a meta-gradient.

\cite{CountBasedExplo} also addressed unsupervised RL in multiple environments concurrently with \cite{unsup}. While their setting is similar, they proposed a fundamentally different solution. They utilized a pre-training objective inspired by count-based methods \cite{CountBasedExplo} instead of our entropy objective. Although they designed a specific bonus for multiple-environment settings, their approach established a uniform preference over the class, prioritizing the worst-case environment differently from \cite{unsup}.

By synthesizing ideas from \cite{CuriousExplo}, which emphasizes a curiosity-based objective, and \cite{unsup}, which prioritizes a state visitation entropy-based objective, we aim to create a more robust framework. While \cite{CuriousExplo} is confined to single-environment scenarios and \cite{unsup} is primarily oriented toward long-term goals with limited focus on shorter-term exploration, especially in common environments, our approach seeks to overcome these shortcomings.

Our strategy involves integrating the best aspects of both approaches to enhance their synergy. By doing so, we strive to establish a more versatile and adaptive system that not only addresses the limitations of each individual method but also leverages their respective strengths. This integration aims to create a unified approach capable of balancing short-term exploration and long-term learning across diverse and challenging environments.

    
\section{Improvements}

\subsection{Sampling Trajectories Using a PDF}
Instead of just using the worst-performing trajectories to learn an exploration policy as done in \cite{unsup}, we construct a probability distribution over the entropies of all the trajectories. The distribution is designed such that the trajectories with the worst/lowest entropies have the highest probabilities and vice-versa. We achieve this by computing the probability of the $i^{th}$ trajectory as:

\begin{align}
    p_i = softmax(-e_i)
\end{align}
where $e_i$ is the state visitation entropy of the $i^{th}$ trajectory

Then, we simply sample a fixed number of trajectories from this distribution and use these sampled trajectories to learn the exploration policy. Thus, the policy will learn not only from the worst-performing trajectories (although they will constitute the majority of the sampled trajectories) but also from a few easier/good trajectories. The intuition behind this approach was that sampling from a probability distribution would introduce some stochasticity and allow the policy to generalize better by learning from both, easy and difficult trajectories.

\subsection{Dynamic Alpha}
First, we will define $\alpha$ in this scenario.

Consider a random variable $X$ with a cumulative density function (CDF) as follows:
\begin{equation}
    F_X(x) = Pr(X \le x)
    \label{eqn:cdf}
\end{equation}
We can define the Value-at-Risk (VaR) as follows:
\begin{equation}
    VaR_\alpha(X) = inf \{x| F_X(x) \le \alpha\}
    \label{eqn:VaR}
\end{equation}
where $\alpha$ is our confidence level and the VaR is essentially the $\alpha$-percentile.

In \cite{unsup}, the objective function is defined as the Conditional Value-at-Risk (CVaR), i.e., the mean of the $\alpha$-percentile entropies as below:
\begin{align}
    CVaR_\alpha(X) = E[X | X \leq VaR_\alpha(X)]
\end{align}

Thus, when $\alpha$ is lower, the lower the probability is of a bad exploration outcome after training for a given environment from a class of Controlled Markov Processes as we hedge more against such outcomes. In the original paper, \cite{unsup}, the confidence level ($\alpha$) was kept constant throughout training. We attempt to improve overall results by dynamically changing $\alpha$ during training. We tried several variations of this idea such as increasing $\alpha$ over time, using different scheduling techniques, and decreasing $\alpha$ over time. We obtained the best results when the percentile was decreased by 1 every $x$ epochs where: 
\[x = \left \lfloor \dfrac{\text{Total number of epochs}}{\text{Total number of batches - Input percentile}}  \right \rfloor \]
That new percentile is then used for $\alpha$, where $\alpha$ is $ percentile / num\_batches$.

This makes sense because as training progresses, we should keep decreasing the $\alpha$ value so that our model improves more over time by lowering its tolerance for "bad outcomes". Essentially, as our model learns, it will do better for the easier trajectories and we want it to continue learning to perform better for harder and harder trajectories.

\subsection{Higher KL Divergence Threshold}
Another approach we tried to improve performance was to increase the KL Divergence threshold. In the original paper, they use a threshold on the KL Divergence between the policy from the previous epoch (behavioral policy) and the latest/updated policy (target policy) as a stopping condition for the number of off-policy optimization steps taken within an epoch of training. The behavioral policy is used to collect trajectories at the beginning of each epoch and the off-policy optimizations are done using this collected data within the epoch. Thus, by increasing the threshold, the target policy is allowed to diverge more from the behavioral policy during the off-policy optimization stage. 

The original KL Divergence threshold may have been too restrictive, constraining the policy to not explore the environment freely. Typically, the KL divergence is used in Trust Region Policy Optimization (TRPO) to make learning more conservative and to prevent overshooting over the local minima during optimization. This is achieved by preventing the learnt policy from diverging too much from a fixed starting policy. The lower the threshold for the KL Divergence is, the more conservative the learning will be. However, such a conservative approach may be excessive for the pretraining phase where there is no extrinsic reward from the environment and only an intrinsic reward. Our intuition is that during this phase it is better to learn a more robust policy that can explore adequately and thus it is better to allow a more free learning process. 

\subsection{Curiosity-Driven Exploration}
When someone is curious, they have a desire to know something that they don't currently know. In RL, \cite{CuriousExplo} proposed an analogous definition of curiosity for an agent as the error in the agent’s ability to predict the consequences of its own actions. The more uncertain an agent is about what will occur after taking some action, the more curious it is said to be. If an agent can accurately predict what will happen given an action, the agent is less curious. We use curiosity to try to improve the performance of the method from \cite{unsup}. The idea behind this approach is that if the agent is more unsure of the outcome of taking an action in a given state, then the agent has not explored the current space enough and needs to do so.

Curiosity-driven exploration is implemented as described in \cite{CuriousExplo}. For the sake of brevity, we will simply summarize the method here, please see \cite{CuriousExplo} for more details. There are two components to this approach. The first involves a forward dynamics model that tries to predict the next state given the current state and the action taken in the current state as input. The error in this model's prediction is defined as the curiosity term. The policy is trained to maximize this curiosity term while the forward dynamics model is optimized to minimize the error in its predictions. The 2nd component termed the inverse dynamics model tries to predict the action taken to make the transition between two given consecutive states. The purpose of this component is to learn a mapping from a higher dimensional state space to a lower dimensional feature space that encodes all the essential state information. For our experiments, due to the constraints of limited compute and time, we were only able to train and test on relatively simpler environments with very low dimensional observation spaces. Thus, for the curiosity experiment, we only implemented and used the forward dynamics model. We do not learn a state encoder or an inverse dynamics model due to the low dimensional observation space.

\subsection{$\alpha$-Percentile Sampling Over Curiosity}
Expanding upon the preceding discussion, our approach here involves leveraging the $\alpha$-percentile most curious trajectories to facilitate the learning process of the policy. This strategic choice is driven by the intention to compel the policy to further explore the trajectories it is most uncertain about as these trajectories probably cover states that have not been explored sufficiently by the policy. By selecting trajectories within the $\alpha$-percentile highest curiosity scores, we encourage the model to focus on and learn from instances that may initially seem less explored. This deliberate optimization of the policy on only the most uncertain trajectories is motivated by the belief that understanding and effectively navigating challenging and unknown scenarios is vital in enhancing the overall robustness and adaptability of the agent.

\section{Results}
We will now discuss the results of our most significant experiments below. Note that we show results primarily on the grid world with slope class of environments from \cite{unsup} which consists of two environments, an environment with a north-facing slope and the other with a south-facing slope. We also show a few results on another class of environments of an ant navigating a set of stairs. 

\subsection{Unsupervised Pre-Training Results}

\subsubsection{Grid World Environment Pre-Training}

In the Grid World environment, the agent is located in a space with 4 square "rooms" connected as a loop through 4 narrow "hallways". The shape of Grid World can be seen in the heat maps below. The space has either a north slope or a south slope. The task for the agent is to travel through the rooms and corridors, either up the slope or down the slope, to reach a goal point with specified x and y coordinates.




\begin{figure}[h]
\centering
 \makebox[\textwidth]{
\begin{subfigure}{0.4\textwidth}
  \centering
  \includegraphics[scale=0.3]{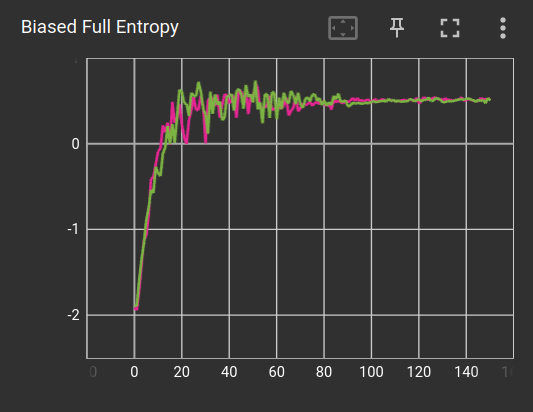}
  \caption{Higher KL}
  \label{kl_entropy}
\end{subfigure}
\begin{subfigure}{0.4\textwidth}
  \centering
  \includegraphics[scale=0.3]{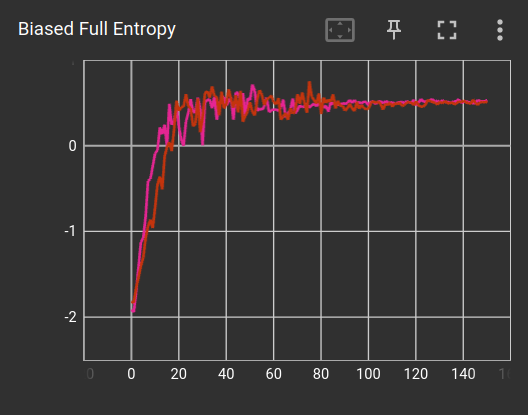}
  \caption{Decreasing $\alpha$}
  \label{dec_alpha_entropy}
\end{subfigure}
\begin{subfigure}{0.4\textwidth}
  \centering
  \includegraphics[scale=0.32]{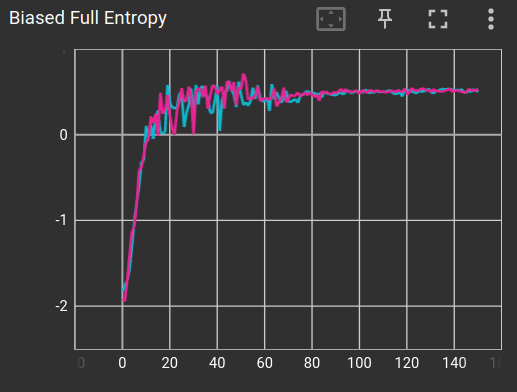}
  \caption{Curiosity}
  \label{curiosity_entropy}
\end{subfigure}
}

\makebox[\textwidth]{
\begin{subfigure}{0.4\textwidth}
  \centering
  \includegraphics[scale=0.3]{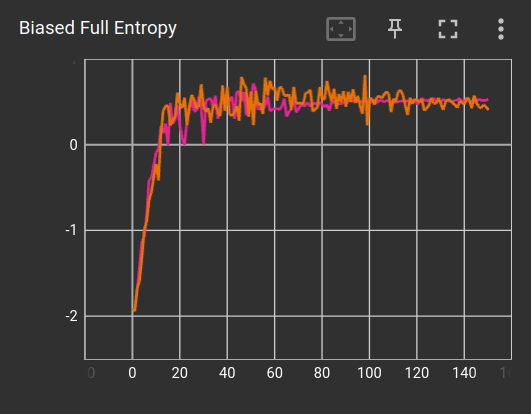}
  \caption{Decreasing $\alpha$ + Higher KL}
  \label{dec_alpha_kl_entropy}
\end{subfigure}
\begin{subfigure}{0.4\textwidth}
  \centering
  \includegraphics[scale=0.31]{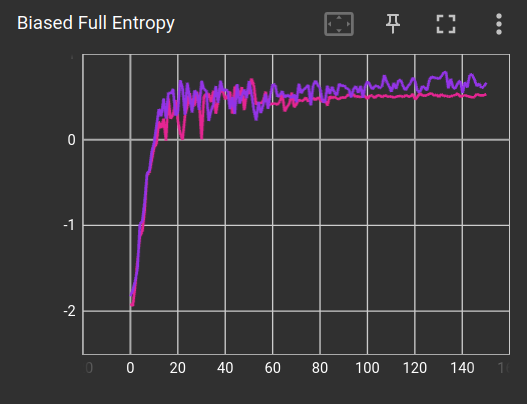}
  \caption{Curiosity + Higher KL}
  \label{curiosity_kl_entropy}
\end{subfigure}
}

\caption{Grid World Entropy Comparison Plots - In all of the above plots, the pink line/plot corresponds to the baseline model.}
\label{grid_entropy}
\end{figure}

We plot the mean entropy over all the collected trajectories at the end of each epoch in Figure \ref{grid_entropy} for our various experiments. From these plots, we clearly see that of all our experiments, only the experiment with the additional curiosity term and the higher KL Divergence threshold together outperforms the baseline and results in an exploration policy that returns higher entropy trajectories.

\begin{figure}[H]
\centering
\makebox[\textwidth]{
\begin{subfigure}{.5\textwidth}
  \centering
  \includegraphics[scale=0.2]{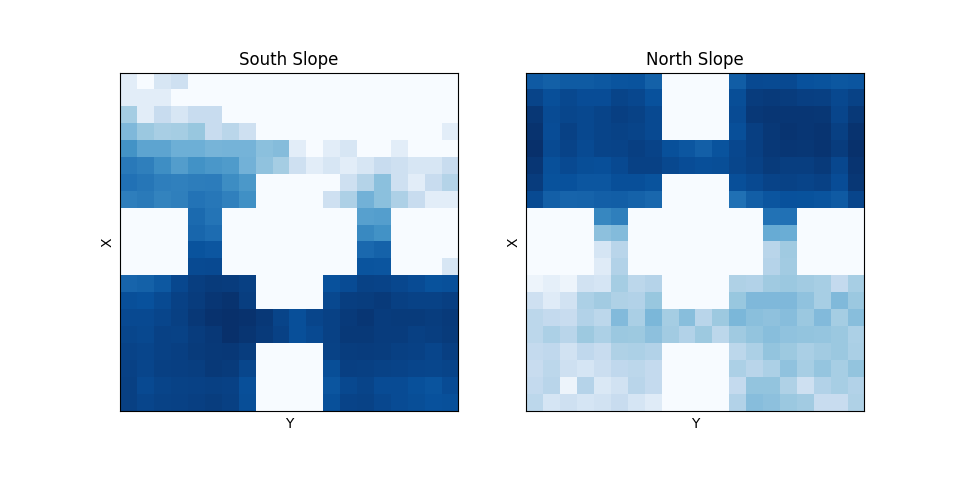}
  \caption{Baseline}
  \label{baseline_heat}
\end{subfigure}
\begin{subfigure}{.5\textwidth}
  \centering
  \includegraphics[scale=0.2]{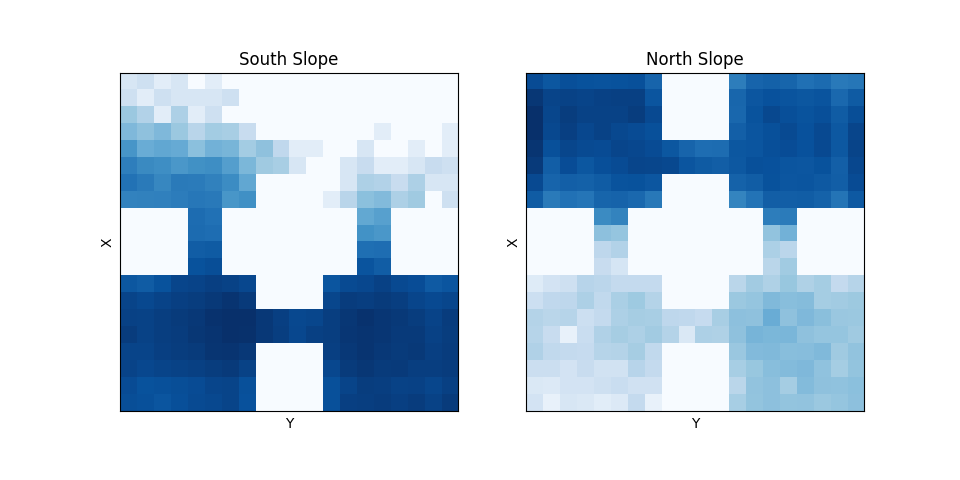}
  \caption{Higher KL}
  \label{kl_heat}
\end{subfigure}
\begin{subfigure}{.5\textwidth}
  \centering
  \includegraphics[scale=0.2]{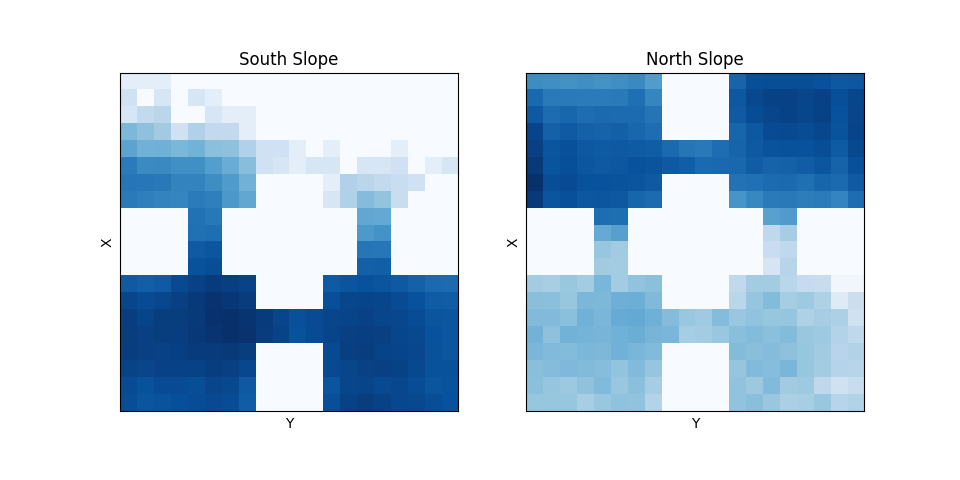}
  \caption{Decreasing $\alpha$}
  \label{dec_alpha_heat}
\end{subfigure}
}
\newline
\makebox[\textwidth]{
\begin{subfigure}{.5\textwidth}
  \centering
  \includegraphics[scale=0.2]{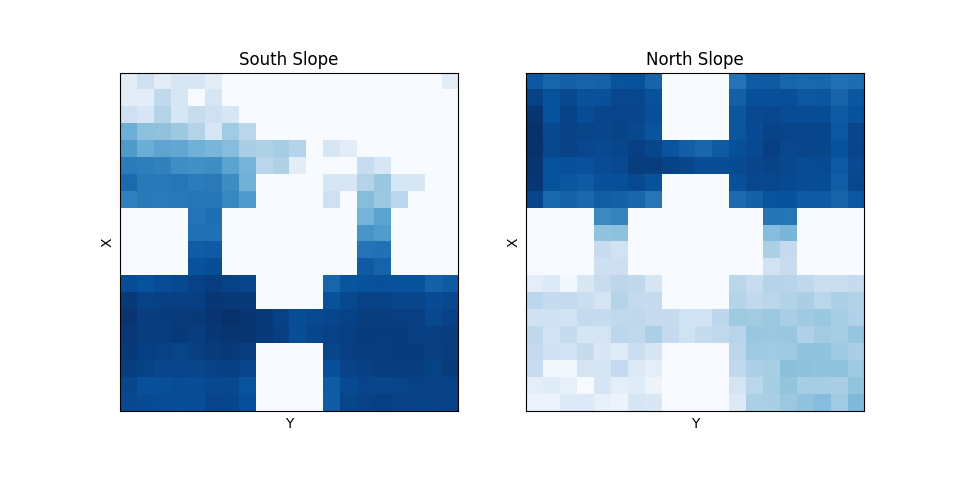}
  \caption{Curiosity}
  \label{curiosity_heat}
\end{subfigure}
\begin{subfigure}{.5\textwidth}
  \centering
  \includegraphics[scale=0.2]{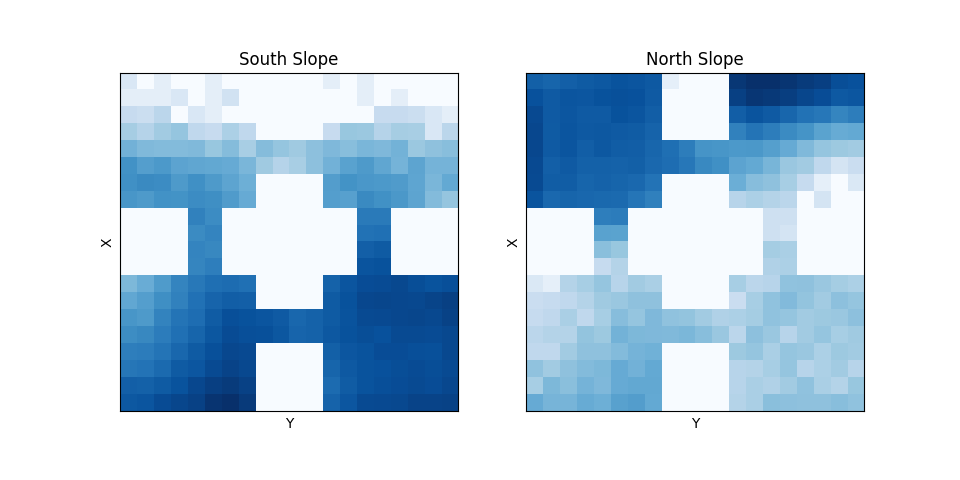}
  \caption{Decreasing $\alpha$ + Higher KL}
  \label{dec_alpha_kl_heat}
\end{subfigure}
\begin{subfigure}{.5\textwidth}
  \centering
  \includegraphics[scale=0.2]{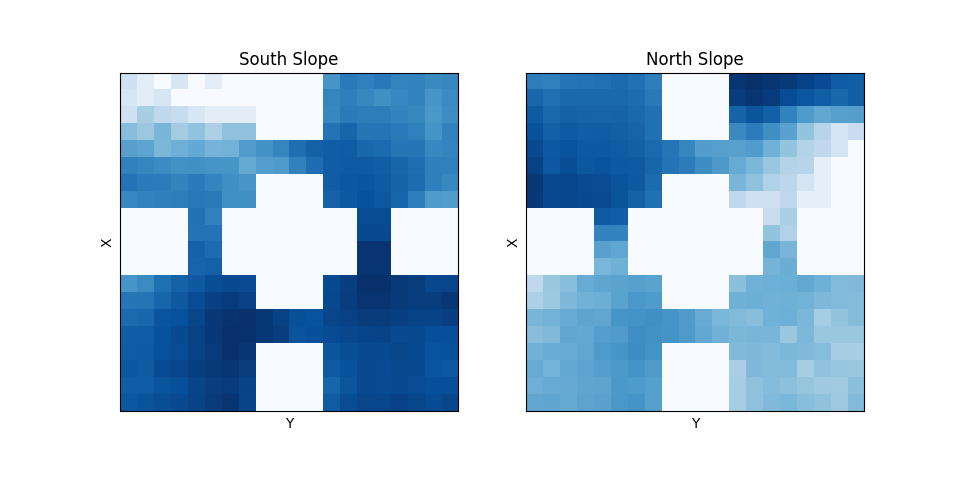}
  \caption{Curiosity + Higher KL}
  \label{curiosity_kl_heat}
\end{subfigure}
}
\caption{Grid World Exploration Heatmaps}
\label{grid_heatmap}
\end{figure}

In Figure \ref{grid_heatmap}, we plot the exploration heatmaps for the class of environments we trained on for our various experiments. These heatmaps indicate how often each state/region in the environment has been visited by the agent with a darker blue indicating more frequent visits. Ideally, the entire environment's map should be shaded in a dark blue as it would indicate that the agent almost equally visits all states/regions in the map. We see from the plots that as expected, the best results are obtained from the experiment with the combination of the additional curiosity term and the higher KL Divergence threshold. However, we also observe that our decreasing $\alpha$ and the combination of decreasing $\alpha$ and higher KL Divergence threshold experiments, both beat the baseline (but not our best experiment) in exploration even though there was no significant improvement observed in the quantitative results covered via the entropy plots above.

Although not shown here, we obtained similar performance to the baseline for our sampling trajectories using a PDF experiment. We suspect that this did not work as expected probably due to the relatively small sampling space where the number of samples is quite small not allowing a sufficient number of easy trajectories to be sampled alongside the tougher ones resulting in a lack of diverse samples. Our experiment of using the $\alpha$-percentile most curious trajectories to optimize the policy did not perform as expected as it gave a similar performance to the baseline. We are currently unsure of the reason behind this and will dive deeper to root out the cause.

\subsubsection{Ant Environment Pre-Training}

In this environment, the ant agent is located in a 3D environment enclosed within 4 walls. The enclosed area is a flight of stairs. The task for the ant agent is to go up or down the stairs to reach a finishing line with a certain distance from the starting line.

\begin{figure}[h]
\centering

\begin{subfigure}{0.4\textwidth}
  \centering
  \includegraphics[scale=0.3]{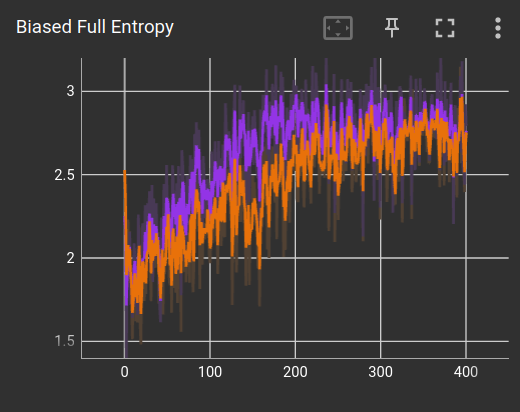}
  \caption{Higher KL}
  \label{kl_entropy}
\end{subfigure}
\begin{subfigure}{0.4\textwidth}
  \centering
  \includegraphics[scale=0.31]{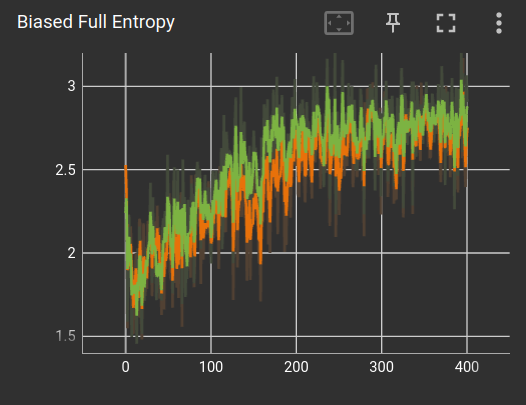}
  \caption{Curiosity + Decreasing $\alpha$ + Higher KL}
  \label{curiosity_dec_alpha_kl_entropy}
\end{subfigure}

\caption{Ant Entropy Comparison Plots - In all of the above plots, the orange line/plot corresponds to the baseline model.}
\label{ant_entropy}
\end{figure}

Similar to Grid World, we observe that our experiment with the combination of curiosity, decreasing $\alpha$ and higher KL Divergence threshold performs the best resulting in an exploration policy that returns higher entropy trajectories. We also see that just the higher KL Divergence threshold experiment performs better than the baseline.

\begin{figure}[H]
\centering
\makebox[\textwidth]{
\begin{subfigure}{.5\textwidth}
  \centering
  \includegraphics[scale=0.2]{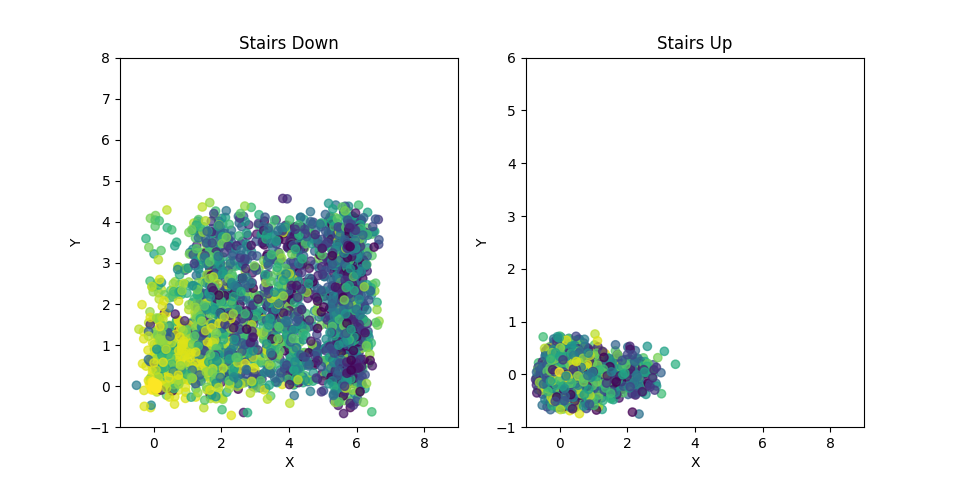}
  \caption{Baseline}
  \label{baseline_heat}
\end{subfigure}
\begin{subfigure}{.5\textwidth}
  \centering
  \includegraphics[scale=0.2]{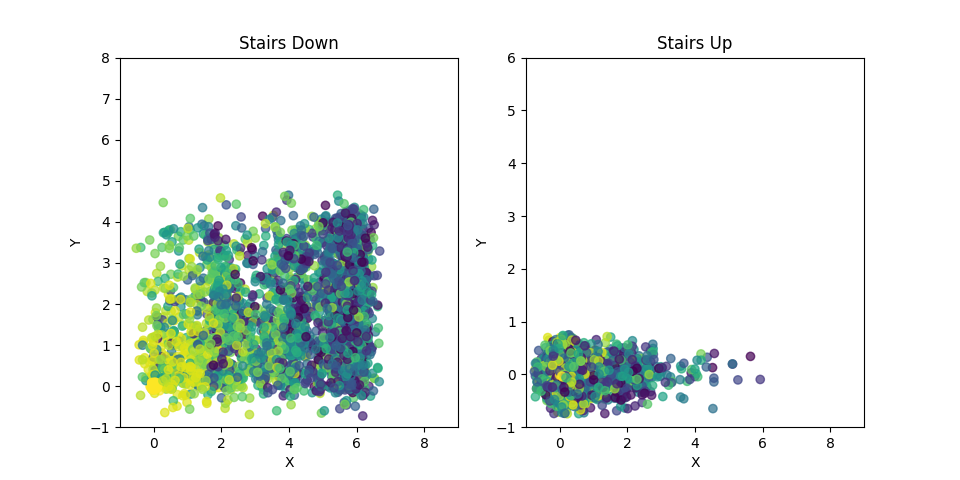}
  \caption{Higher KL}
  \label{kl_heat}
\end{subfigure}
\begin{subfigure}{.5\textwidth}
  \centering
  \includegraphics[scale=0.2]{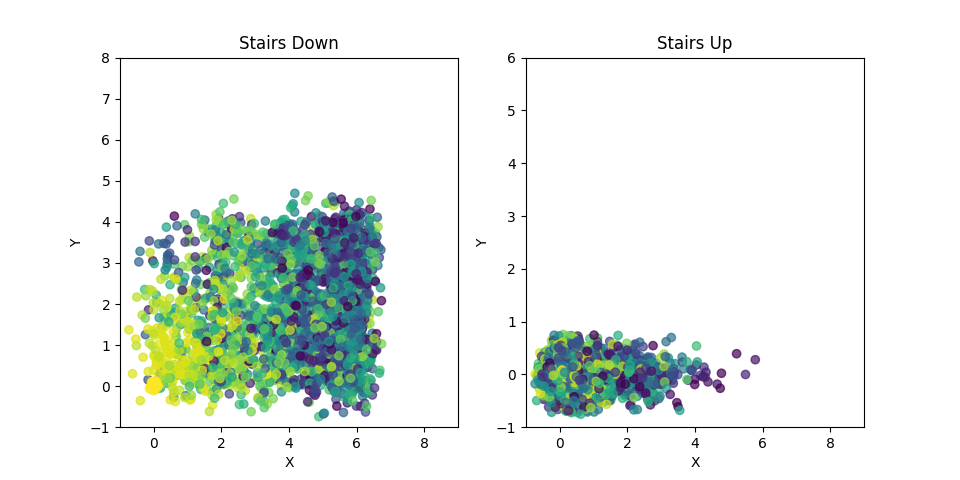}
  \caption{Curiosity + Decreasing $\alpha$ + Higher KL}
  \label{curiosity_alpha_dec_kl_heat}
\end{subfigure}
}
\caption{Ant Exploration Heatmaps}
\label{ant_heatmap}
\end{figure}

We see similar trends in performance in the exploration heatmaps in Figure \ref{ant_heatmap} as was seen in the entropy plots in Figure \ref{ant_entropy}. We see that the point cloud is less concentrated and more spread out for both our experiments with the spread being slightly greater for our best experiment. 

\subsection{Supervised Fine Turned Results}

\subsubsection{Grid World Environment Fine-tuning}

For the Grid World environment, the supervised fine tuning was done using randomly generated goal points. The agent was trained using 5 different goals over a total of 400 epochs. Two frames of the process can be seen below. The green dot indicates the goal position while the blue dot is the agent.

\begin{figure}[H]
\centering

\begin{subfigure}{0.4\textwidth}
  \centering
  \includegraphics[scale=0.2]{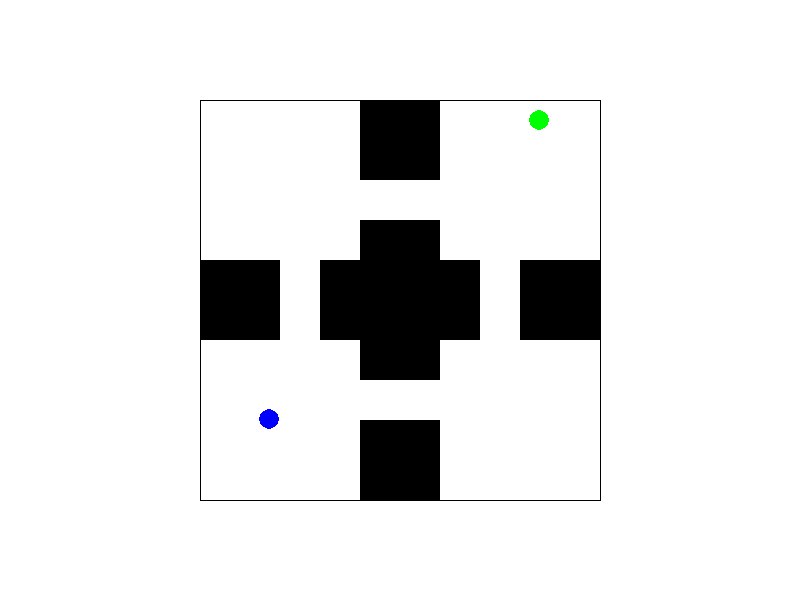}
  \caption{Agent at starting position}
  \label{kl_entropy}
\end{subfigure}
\begin{subfigure}{0.4\textwidth}
  \centering
  \includegraphics[scale=0.2]{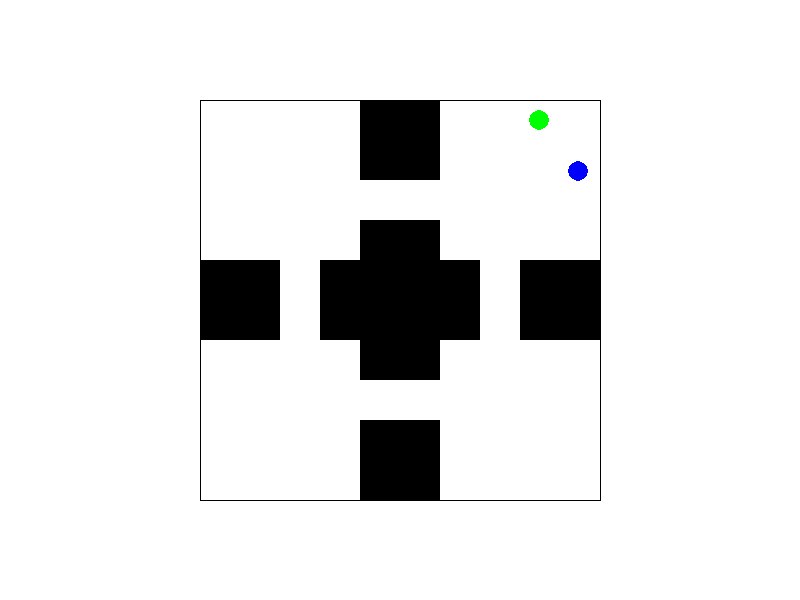}
  \caption{Agent approaching the goal}
  \label{curiosity_dec_alpha_kl_entropy}
\end{subfigure}

\caption{Output frames of our model's fine-tuning process in Grid World}
\label{grid_sup}
\end{figure}

We use the same random seed for all our experiments to ensure that we are finetuning and comparing results for the same goals. The average return shows how well the agent is able to reach the goal in the figure below:

\begin{center}
\includegraphics[scale=0.2]{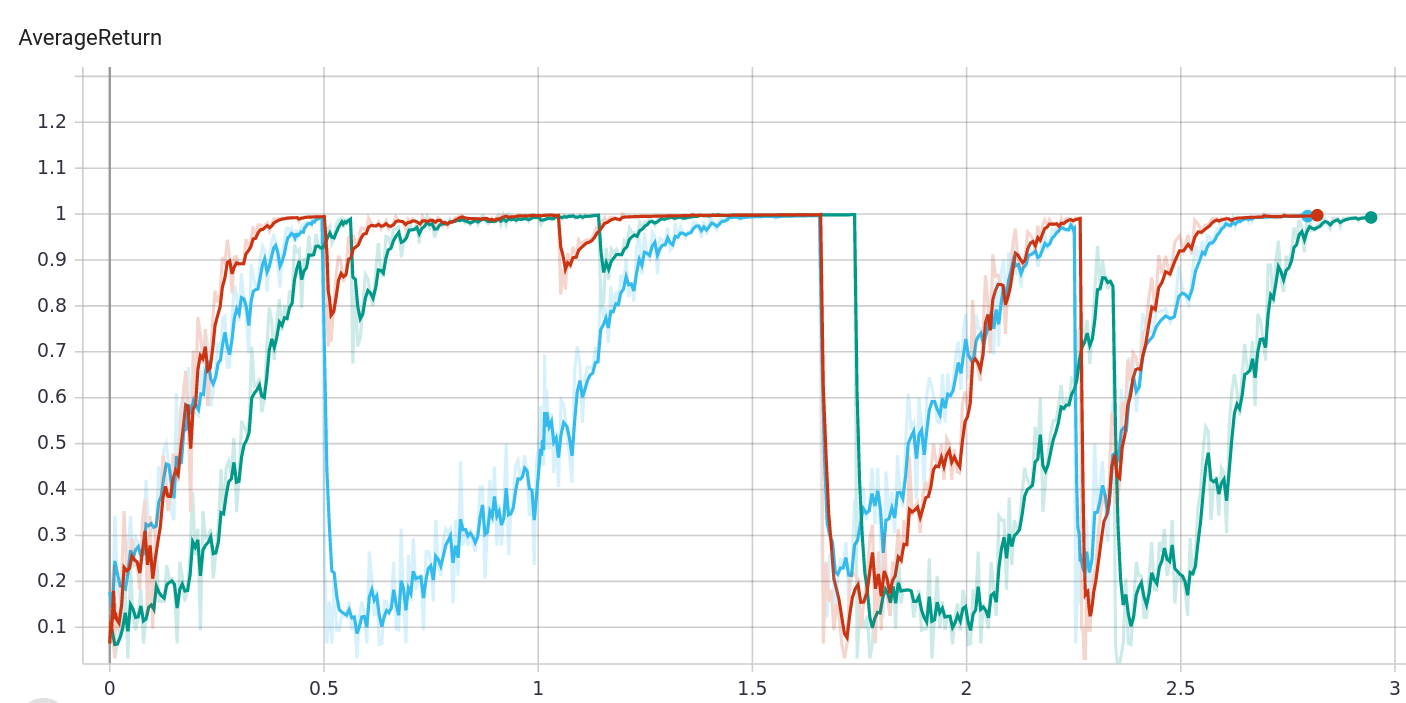}
\end{center}

The blue curve corresponds to our best model pre-trained with the additional curiosity term and a higher KL Divergence threshold of 100, the green curve indicates the model pre-trained only with a higher KL Divergence threshold of 100, and the red curve indicates the baseline model. The curves drop each time the goal is changed. Our model performs the best for goals 1, 4, and 5 while the baseline does better for goals 2 and 3.

\subsubsection{Ant Environment Fine-tuning}
For the Ant environment, the fine-tuning was done for 8 different goals, each further away from the starting line than the one before making it increasingly difficult for the ant to reach the finishing line. The ant needs to move up or down the stairs to reach the 8 finishing lines. 

\begin{figure}[h]
\centering

\begin{subfigure}{0.4\textwidth}
  \centering
  \includegraphics[scale=0.3]{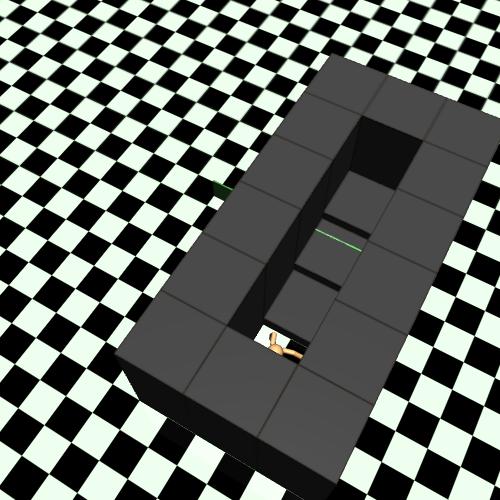}
  \caption{Ant at starting position}
  \label{kl_entropy}
\end{subfigure}
\begin{subfigure}{0.4\textwidth}
  \centering
  \includegraphics[scale=0.3]{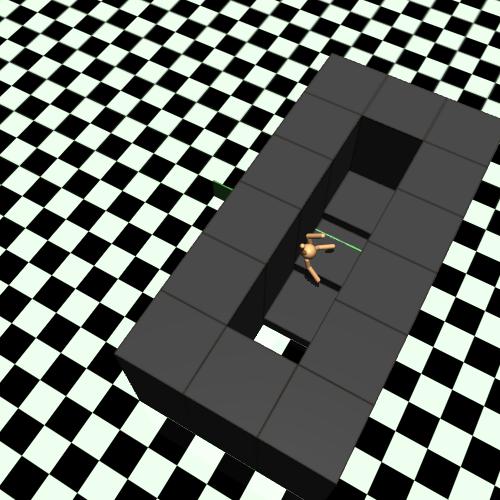}
  \caption{Ant reaching the goal}
  \label{curiosity_dec_alpha_kl_entropy}
\end{subfigure}

\caption{Output frames of our model's fine-tuning process in the Ant environment}
\label{ant_sup}
\end{figure}

In Figure \ref{ant_sup}, the green line indicates the finishing line (goal). 
\begin{figure}
\begin{center}
\includegraphics[scale=0.3]{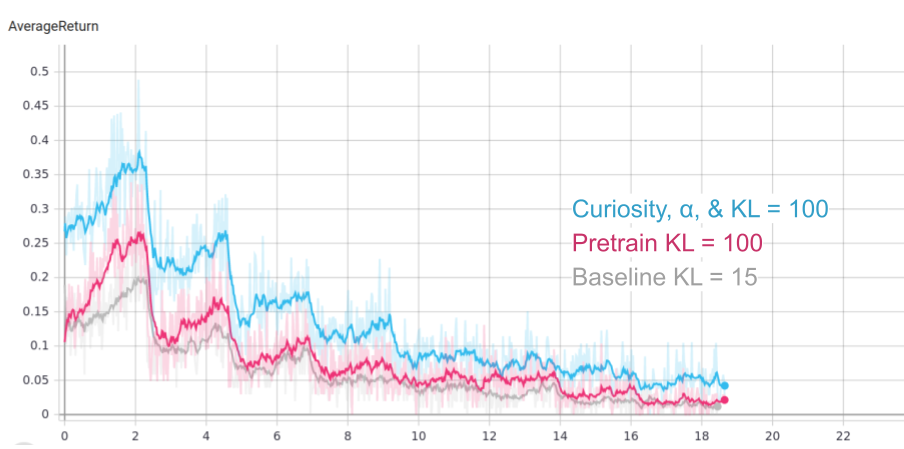}
\caption{The average return for 3 experiments. The blue curve indicates our best model pre-trained with the curiosity term, decreasing $\alpha$, and a higher KL Divergence threshold of 100. The pink curve indicates the model pre-trained only with a higher KL Divergence threshold of 100. The grey curve indicates the baseline model. }
 \label{returns}
\end{center}
\end{figure}

Since the 8 goals get increasingly challenging, we observe a decreasing trend in the return curves in general. However, within the training epochs of each goal, we still observe a visible increase in average return as displayed in figure \ref{returns}. The sharp drops in the curves are due to the goals being changed, i.e., the finishing line moving further away. A higher return indicates that the ant reaches the finishing line more frequently. 

We clearly observe that our model has a significantly better performance than the other models in the Ant environment. The improvement is more obvious here because the ant has more joints and this environment is much more challenging and complex. Thus, we see the promise of our proposed approach of using a curiosity term, decreasing $\alpha$, and using a higher threshold for the KL Divergence during the pre-training phase to get more a robust policy especially when dealing with more challenging environments.

\subsection{Analysis}

We observe that introducing curiosity drastically improved performance for a more complex environment like Ant. It did not show its full potential in simpler environments like Grid World as the model can not be curious enough in such environments where it can easily learn and know the consequences of its actions. Just increasing the KL Divergence threshold also improved the performance more visibly in the Ant environment. Finally, decreasing $\alpha$ over time allows us to learn from worse and worse trajectories as the model gets better over time and has shown positive effects to a certain extent.



\section{Conclusion}
\label{sec:conclusion}

The integration of curiosity-driven exploration, an increased KL Divergence threshold, and a decreasing $\alpha$ over time has proven to enhance performance. To further advance this research, exploring more complex environments is essential, particularly to experiment with incorporating a learned inverse dynamics model as well. Increasing the number of sampled trajectories would offer a more comprehensive evaluation of using our probability distribution sampling approach for improved performance. Additionally, future research directions could involve dynamically learning the KL-Divergence threshold and adapting it to the evolving characteristics of the environment to learn more flexible and efficient exploration strategies.




\bibliography{main} 

\end{document}